%% file: main.tex
\documentclass{article}
\usepackage{caption}

\usepackage[preprint]{corl_2026}

\usepackage{amsmath, amssymb, amsfonts}
\usepackage{enumitem}
\usepackage{booktabs}
\usepackage{graphicx}
\usepackage{xcolor}
\usepackage{microtype}
\usepackage{tikz}
\usetikzlibrary{positioning, arrows.meta, shapes.geometric, fit, backgrounds, calc}

\title{AnyGoal: Vision-Language Guided Multi-Agent Exploration
       for Training-Free Lifelong Navigation}

\author{
MoniJesu James \and
    Marcelino Julio Fernando \and
        Miguel Altamirano Cabrera \and
            Dzmitry Tsetserukou
}

\begin{document}
\maketitle

\begin{abstract}
End-to-end navigation policies trained on large simulation corpora degrade
sharply when transferred to scenes, target categories, or goal modalities
outside their training distribution. The leading zero-shot alternatives
inherit a different but equally constraining ceiling: modular pipelines
such as Modular~GOAT are bottlenecked by the recall of a closed-set
object detector, while recent 3D snapshot--memory systems
(e.g.\ 3D-Mem) accumulate dense, view-dependent representations that are
heavy to maintain and difficult to share across embodiments. We present
\textbf{AnyGoal}, a fully decentralized, training-free multi-robot
architecture that places a Vision--Language Model (VLM) at the cognitive
core of frontier-based exploration and coordinates the swarm through a
shared 2D \emph{Gaussian Bayesian Value Map} (BVM). The BVM maintains a
per-pixel $(\mu, \sigma^2)$ posterior over goal relevance, updated by
precision-weighted fusion of VLM scores through a depth-cone mask, and
is \emph{never reset between subtasks}, yielding lifelong evidence
accumulation. Frontiers are ranked by a convex blend of a VLM-as-judge
softmax and a Bayesian Upper-Confidence-Bound (UCB) term computed
directly on the BVM. A sequential greedy allocator with a
spatial-separation penalty and commitment hysteresis distributes
frontiers across agents without any centralized controller. On the full
\textsc{GOAT-Bench} \emph{val\_unseen} split (360 episodes,
2{,}669~subtasks), our dual-agent system achieves \textbf{52.4\,\%}
Subtask~SR at \textbf{12.7\,\%} SPL, state-of-the-art under the
strict physical regime (discrete $0.25\,\mathrm{m}$ steps, no
teleportation, $42^{\circ}$ HFOV) and a \textbf{+27.5~percentage-point}
improvement over the published Modular~GOAT baseline (24.9\,\%). The
single-agent AnyGoal configuration alone achieves
\textbf{41.9\,\%} Subtask~SR, demonstrating that the gains arise from
the underlying decision architecture rather than from raw agent count.
A controlled four-way perception ablation shows that
\emph{high-recall open-vocabulary detectors shift the dominant failure
mode of zero-shot navigation away from exploration and onto goal
verification at the stopping decision}.
\end{abstract}


\section{Introduction}
\label{sec:intro}

Embodied agents that operate in real homes will be asked, in a single
deployment, to find an object by name, by photograph, and by free-form
natural language description, all while persisting useful memory across
tasks. \textsc{GOAT-Bench}~\citep{khanna2024goat} crystallized this
requirement as \emph{multi-modal lifelong navigation}: an agent receives a
stream of $5$--$10$ heterogeneous subtasks per episode in photorealistic
HM3D scenes~\citep{ramakrishnan2021habitat}, with no scene or category
supervision at training time. Two regimes have emerged on this benchmark.
End-to-end policies, most prominently SenseAct-NN Skill
Chain~\citep{khanna2024goat} and the concurrent
AstraNav-Memory~\citep{ren2025astranav}, attain strong path efficiency
on familiar distributions, but their reliance on learned spatial priors
makes them brittle under category, modality, and embodiment shift. Modular zero-shot
pipelines such as Modular~GOAT~\citep{khanna2024goat} side-step training by
composing off-the-shelf detection, mapping, and planning stacks, but inherit
the \emph{closed-vocabulary} ceiling of detectors such as
DETIC~\citep{zhou2022detecting}; on \textsc{GOAT-Bench} \emph{val\_unseen} the
strongest Modular~GOAT variant attains $24.9\,\%$ Subtask~SR and
$17.2\,\%$~SPL. A more recent line of work, exemplified by
3D-Mem~\citep{yang20253d}, replaces the implicit detector with a
view-dependent 3D snapshot database queried by a VLM. This restores
open-vocabulary recall at the cost of substantial memory and compute
overhead, and remains structurally single-agent.

We argue that neither extreme is necessary. The \emph{spatial} reasoning
that modular systems delegate to fragile detectors and the \emph{semantic}
reasoning that 3D systems delegate to per-snapshot VLM calls can be unified
inside a single lightweight 2D structure, a \emph{Gaussian Bayesian Value
Map}, if the VLM is treated as a noisy oracle whose answers are fused into
a persistent spatial posterior. Once frontiers carry calibrated
$(\mu, \sigma^2)$ belief, they admit a principled blend of
\emph{exploitation} (VLM relevance) and \emph{exploration} (UCB on the BVM),
and they become a natural medium for decentralized multi-agent allocation:
each agent reads the same map but acts greedily under a separation penalty
that no centralized scheduler has to enforce.

\paragraph{Contributions.}
This paper introduces \textbf{AnyGoal}, a fully decentralized, training-free
multi-robot architecture for multi-modal lifelong navigation, and validates
it through the most comprehensive perception ablation reported on
\textsc{GOAT-Bench} \emph{val\_unseen} to date. Concretely:

\begin{enumerate}[leftmargin=1.4em,topsep=2pt,itemsep=1pt]
\item \textbf{A Gaussian Bayesian Value Map for lifelong relevance.}
We replace the Beta-Bernoulli and binary occupancy heuristics common in
zero-shot navigation~\citep{yokoyama2024vlfm,gadre2023cows} with a
precision-weighted Gaussian update over per-pixel $(\mu, \sigma^2)$
goal-relevance belief, fused through a depth-cone observation mask. The map
is never reset between subtasks, supporting cross-subtask evidence
carry-over and natural hallucination decay.
\item \textbf{Convex-blended VLM-as-judge with Bayesian UCB.}
Candidate frontiers are rendered as A/B/C/D markers on the egocentric view;
the VLM returns a softmax. We blend the softmax with a UCB term on the
BVM via a convex coefficient ($w\!=\!0.50$), gated by an explicit
``is-worth'' Yes/No prompt~($\geq\!0.50$). This recovers the ergodic
exploration guarantees of UCB while letting the VLM dominate when its
preference is sharp.
\item \textbf{Decentralized swarm coordination without explicit messaging.}
Frontier allocation is a sequential greedy bid over the shared BVM with a
spatial-separation penalty
($d\!<\!100\,\mathrm{px}\!\Rightarrow\!\times\!0.6$) and a
$1.5\!\times$ commitment hysteresis. No centralized controller and no
explicit inter-agent message channel are required beyond the shared map.
\item \textbf{Headline results on \textsc{GOAT-Bench}}.
On the full \emph{val\_unseen} split (360 episodes, 2{,}669 subtasks),
AnyGoal with $N\!=\!2$ ground agents achieves \textbf{52.42\,\%} subtask~SR
and \textbf{12.66\,\%} SPL, exceeding Modular~GOAT by \textbf{+27.5~pp}
and the strongest zero-shot single-agent system TANGO by \textbf{+20.3~pp}
on subtask~SR. Our $N\!=\!1$ baseline already attains \textbf{42.0\,\%}
subtask~SR, demonstrating that the architectural prior, not raw agent
count, is doing most of the work.
\item \textbf{A scientific characterization of the perception bottleneck.}
A four-way ablation across SAM3, SAM3+DINOv2, GroundedSAM, and
YOLO-World+MobileSAM, with detector quality varied and every other
component held fixed, shows that the dominant failure mode of zero-shot
navigation \emph{shifts} as recall increases: high-recall SAM3 fails
through premature \textsc{stop} calls ($80.2\,\%$ of failures), while
low-recall YOLO-World exhausts its step budget ($42.3\,\%$). This
reframes the open problem from ``find better detectors'' to ``verify the
goal at the stopping decision''.
\end{enumerate}

The remainder of the paper develops these claims.
Section~\ref{sec:related} positions AnyGoal against the closed-set
modular line, the 3D snapshot-memory line, and centralized multi-agent
reinforcement learning; Section~\ref{sec:method} details the BVM, the
convex-blended UCB score, and the decentralized allocator;
Section~\ref{sec:setup} reports the evaluation protocol;
Sections~\ref{sec:results}--\ref{sec:discussion} present the
state-of-the-art result, the multi-agent scaling curve, the four-way
perception ablation, and a discussion that interprets SPL across
simulation regimes; Section~\ref{sec:limitations} states the remaining
limitations.

\section{Related Work}
\label{sec:related}

\paragraph{Multi-modal lifelong navigation and closed-set perception.}
\textsc{GOAT-Bench}~\citep{khanna2024goat} consolidates object-, image-,
and language-goal navigation into a single lifelong protocol on
HM3D~\citep{ramakrishnan2021habitat}. Its reference baselines, Modular~GOAT
(classical SLAM$+$DETIC~\citep{zhou2022detecting}) and SenseAct-NN Skill
Chain, both rely on \emph{closed-set} vocabularies, and the benchmark
attributes their ceiling to detector recall on out-of-distribution
categories. Open-vocabulary backbones
(GroundingDINO~\citep{liu2024grounding}, YOLO-World~\citep{cheng2024yolo}, SAM3~\citep{carion2025sam}) relax this at the
detection level, but systems that bolt them on without a memory of
\emph{where} the goal was last plausible remain susceptible to premature
stops on near-duplicates (\S\ref{subsec:results-ablation}). AnyGoal
back-projects every detection through depth so spatially inconsistent VLM
hits are rejected before reaching the planner, and treats the VLM not as
a categorical oracle but as a noisy posterior fused into a calibrated map.

\paragraph{Memory in embodied AI.}
Navigation memory has moved from semantic occupancy
grids~\citep{chaplot2020object} to 3D snapshot representations.
3D-Mem~\citep{yang20253d} is the strongest recent exponent: a
\emph{per-snapshot} 3D-fragment database re-queried by a VLM at every
decision point ($49.6\,\%$ Episode~SR, $28.8\,\%$ SPL). The representation
is single-agent and view-dependent, and the cost of re-rendering snapshots
through a VLM grows linearly in episode length and quadratically with
agent count. The BVM takes the opposite stance, a 2D per-pixel Gaussian posterior
whose storage is constant in episode length and whose merge across
agents is a closed-form precision-weighted sum, placing AnyGoal in the
direct lineage of value-map navigation~\citep{yokoyama2024vlfm,huang2023visual}
but with \emph{uncertainty} as a first-class quantity that drives both
UCB exploration and the rejection of spatially inconsistent detections.
VLFM~\citep{yokoyama2024vlfm} is the conceptual ancestor of this line
but reports only on single-modality \textsc{ObjectNav} and has no
published \textsc{GOAT-Bench} \emph{val\_unseen} number; we therefore
do not include it as a direct baseline and instead position the Gaussian
BVM as its principled multi-modal, uncertainty-aware extension.

\paragraph{Zero-shot and multi-agent robotics.}
Zero-shot semantic navigation has progressed via VLM-grounded frontier
methods~\citep{yokoyama2024vlfm,gadre2023cows,huang2023visual},
LLM-planned skill chains~\citep{shah2023lm,huang2023visual}, graph
captioning pipelines such as TANGO~\citep{ziliotto2025tango}, and
unified-goal frameworks such as UniGoal~\citep{yin2025unigoal}; trained
counterparts include MTU3D~\citep{zhu2025move}. All are overwhelmingly
single-agent. Multi-agent navigation is dominated by
centralized MARL~\citep{yu2022surprising,rashid2020monotonic} that requires training,
reward shaping, and tight communication assumptions. We instead reduce
coordination to a \emph{decentralized cost-utility bid} over the shared
BVM, in spirit closest to classical decentralized
auctions~\citep{burgard2005coordinated,gerkey2002sold} but operating over
an uncertainty-aware value map rather than a flat information-gain field,
with no explicit messaging beyond map synchronization.

\section{Method}
\label{sec:method}

\begin{figure}[t]
\centering
\resizebox{\textwidth}{!}{%
\begin{tikzpicture}[
  font=\footnotesize, >=Stealth, node distance=5mm,
  agent/.style={rectangle, draw, rounded corners=2pt, fill=black!6,
     minimum height=0.8cm, minimum width=2.3cm, align=center, inner sep=2pt, line width=0.5pt},
  perc/.style={rectangle, draw, rounded corners=2pt, fill=blue!10,
     minimum height=1.5cm, minimum width=2.6cm, align=center, inner sep=2pt, line width=0.6pt},
  mem/.style={rectangle, draw, rounded corners=2pt, fill=orange!16,
     minimum height=1.5cm, minimum width=2.8cm, align=center, inner sep=2pt, line width=0.7pt},
  coord/.style={rectangle, draw, rounded corners=2pt, fill=yellow!22,
     minimum height=1.5cm, minimum width=3.0cm, align=center, inner sep=2pt, line width=0.7pt},
  plan/.style={rectangle, draw, rounded corners=2pt, fill=green!14,
     minimum height=0.8cm, minimum width=2.1cm, align=center, inner sep=2pt, line width=0.5pt},
  chain/.style={rectangle, draw, rounded corners=1pt, fill=yellow!8,
     align=center, inner sep=2pt, font=\scriptsize, line width=0.4pt, minimum height=0.5cm},
  arr/.style={-{Stealth[length=2mm]}, line width=0.7pt},
  thinarr/.style={-{Stealth[length=1.6mm]}, line width=0.5pt, black!55},
]
\node[agent] (ai) {\textbf{Agent $i$}\\ goal $g_t$\\ {\scriptsize name\,/\,image\,/\,desc}\\ $+$ RGB-D};
\node[agent, below=4mm of ai] (aj) {\textbf{Agent $j$}\\ goal $g_t$ $+$ RGB-D};
\node[perc, right=6mm of ai, yshift=-4mm] (perc)
  {\textbf{Perception}\\ {\itshape swappable}\\ SAM3\,/\,GSAM\,/\\ YOLO-W $+$ MobileSAM\\ SpaceOM VLM $+$ CLIP};
\node[mem, right=7mm of perc] (map)
  {\textbf{Shared Mapping}\\ GoalProjector\\ {\scriptsize(depth-cone mask)}\\ \textbf{2D Gaussian BVM}\\ $(\mu,\sigma^2)$\,\,{\scriptsize never reset}};
\node[coord, right=7mm of map] (coord)
  {\textbf{Decentralized}\\ \textbf{Coordination}\\ {\scriptsize$s(f)\!=\!(1\!-\!w)p_{\mathrm{VLM}}\!+\!w\widetilde{\mathrm{UCB}}$}\\ {\scriptsize$\mathrm{UCB}\!=\!\mu\!+\!\beta\sqrt{\sigma^2}$}\\ {\scriptsize greedy bid $+$ sep.\ penalty}};
\node[plan, right=7mm of coord, yshift=5mm] (plan) {\textbf{Planning}\\ FMM $+$ DistGate};
\node[plan, below=4mm of plan] (act) {\textbf{Action}\\ $\{$F, L, R, \textsc{stop}$\}$};
\node[chain, below=10mm of map.south west, anchor=west] (c1) {scene\\caption};
\node[chain, right=3mm of c1] (c2) {room\\prior};
\node[chain, right=3mm of c2] (c3) {is-worth\\gate};
\node[chain, right=3mm of c3] (c4) {ABCD\\softmax};
\node[left=2mm of c1, font=\scriptsize\itshape, align=right] (cl) {VLM-as-\\Judge chain};
\draw[arr] (ai.east) -- ([yshift=3mm]perc.west);
\draw[arr] (aj.east) -- ([yshift=-3mm]perc.west);
\draw[arr] (perc) -- (map);
\draw[arr] (map) -- (coord);
\draw[arr] (coord.east) -| (plan.south);
\draw[arr] (plan) -- (act);
\draw[thinarr] (c1)--(c2); \draw[thinarr] (c2)--(c3); \draw[thinarr] (c3)--(c4);
\draw[thinarr] (c4.east) -| (coord.south);
\begin{scope}[on background layer]
\node[draw=black!55, dashed, rounded corners=3pt, fit=(map)(coord)(c1)(c4),
   inner sep=3mm, fill=blue!3] (sb) {};
\node[font=\scriptsize\itshape, black!70, below=0.5mm of sb.south]
   {shared across all $N$ agents -- no centralized controller};
\end{scope}
\end{tikzpicture}}
\caption{\textbf{The AnyGoal architecture.} A single forward path runs at
every agent: the multi-modal goal token $g_t$ and an RGB-D observation are
processed by a \emph{swappable} open-vocabulary perception stack (SAM3,
GroundedSAM, or YOLO-World, plus a spatial VLM). Detections are
back-projected by the GoalProjector into the shared 2D Gaussian Bayesian
Value Map (Eq.~\eqref{eq:gaussian-update}). The decentralized coordination
layer scores frontiers via a VLM-as-judge softmax convex-blended with a
Bayesian UCB term (Eq.~\eqref{eq:blend}) and allocates them across agents
under a multiplicative separation penalty (Eq.~\eqref{eq:alloc}). The
mapping and coordination blocks (dashed box) are the only shared substrate;
no centralized controller and no explicit inter-agent message channel are
required.}
\label{fig:arch}
\end{figure}
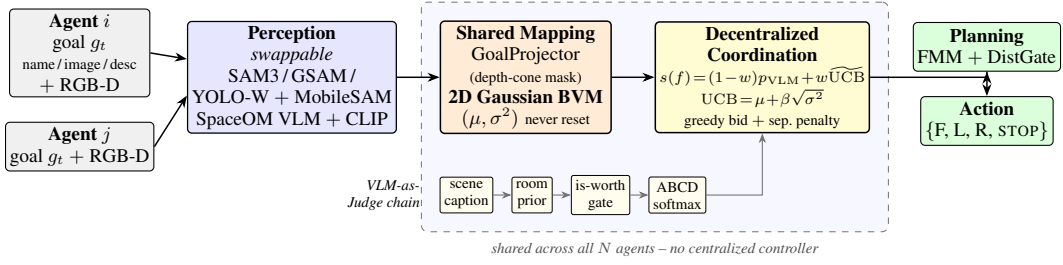

\subsection{System architecture}
\label{subsec:method-arch}

AnyGoal is structured as four \emph{swappable} layers
(Fig.~\ref{fig:arch}). At each decision step, every agent $i$ runs the same
local pipeline:

\begin{enumerate}[leftmargin=1.4em,topsep=2pt,itemsep=1pt]
\item \textbf{Perception.} An open-vocabulary segmentation backbone
(SAM3~\citep{carion2025sam} by default; GroundedSAM~\citep{ren2024grounded}
and YOLO-World~\citep{cheng2024yolo} as ablations) emits
goal-conditioned masks; a spatial VLM (SpaceOM-3B) provides relevance
scores; and for image-goal subtasks an image matcher
(CLIP ViT-B/32~\citep{radford2021learning} by default;
DINOv2~\citep{oquab2023dinov2} as ablation) supplies similarity to the
reference image.
\item \textbf{Mapping.} Detections are back-projected through depth into a
shared BEV occupancy grid at $0.05\,\mathrm{m}/\text{cell}$ by a
\emph{GoalProjector} that rejects spatially inconsistent VLM hits; the
Bayesian Value Map (\S\ref{subsec:method-bvm}) maintains goal-relevance
belief on the same grid.
\item \textbf{Reasoning.} A four-stage prompt chain renders the egocentric
view plus candidate frontiers as A/B/C/D markers; the VLM emits (a) a
scene caption, (b) a room-type prior, (c) an ``is-worth-pursuing'' Yes/No
gate, and (d) a softmax over the frontier markers
(\S\ref{subsec:method-blend}).
\item \textbf{Planning.} Frontier extraction (connected-component
boundaries on the shared BEV), convex-blended scoring, and a
Fast-Marching-Method (FMM) planner~\citep{sethian1996fast} produce a
discrete action in the Habitat action space~\citep{savva2019habitat}.
\end{enumerate}

The same code path runs at every agent: only the shared BEV and BVM act as
implicit communication.

\subsection{Gaussian Bayesian Value Map}
\label{subsec:method-bvm}

Let $\mathcal{M}=\{(x,y)\}$ denote the BEV grid. We attach to each cell
$p$ a scalar belief over the probability that $p$ contains, or is in the
immediate spatial neighborhood of, the current goal:
$\mathrm{Bel}(p)=\mathcal{N}(\mu(p), \sigma^2(p))$. The prior is uniform,
$\mu_0\!=\!0.5,\ \sigma_0^2\!=\!0.5$. At each step every visible cell
receives a new observation
$(\mu_{\text{obs}}, \sigma_{\text{obs}}^2)$, where $\mu_{\text{obs}}$ is
the VLM relevance score projected through the depth-cone visibility mask
and $\sigma_{\text{obs}}^2 = 1 - \mathrm{cone\_conf}(p)$ encodes how
well-supported the projection is by depth. Posterior fusion is the
standard precision-weighted product of Gaussians:
\begin{equation}
\mu_{\text{new}}      \;=\; \frac{\sigma_{\text{obs}}^2\,\mu_{\text{prior}}
                                 + \sigma_{\text{prior}}^2\,\mu_{\text{obs}}}
                                 {\sigma_{\text{prior}}^2 + \sigma_{\text{obs}}^2},
\qquad
\sigma_{\text{new}}^2 \;=\; \frac{\sigma_{\text{prior}}^2 \,\sigma_{\text{obs}}^2}
                                 {\sigma_{\text{prior}}^2 + \sigma_{\text{obs}}^2}.
\label{eq:gaussian-update}
\end{equation}
Eq.~\eqref{eq:gaussian-update} ensures that high-cone-confidence pixels
(sharp depth) dominate the prior while repeated low-confidence observations
decay rather than reinforce an erroneous belief. The BVM is~\emph{never
reset between subtasks}: when the goal token changes, only the VLM scoring
head changes while accumulated geometric and semantic evidence persists,
yielding lifelong evidence carry-over across the $5$--$10$ subtasks per
episode. Across $N$ agents, the swarm posterior is the precision-weighted
fusion of the agent-local beliefs, applied element-wise on the shared
BEV/BVM tensor.

\subsection{VLM-as-judge frontier ranking and convex-blended UCB}
\label{subsec:method-blend}

Let $\mathcal{F}_t$ be the set of frontier cells at step $t$ extracted from
the connected-component boundary of the explored area on the shared BEV.
We rank $\mathcal{F}_t$ by fusing \emph{spatial geometry} with \emph{VLM
common sense}. Up to four salient frontiers are rendered onto the
egocentric image as labeled markers $\{A,B,C,D\}$ and submitted, alongside
the current goal token, to a VLM-as-judge prompt; the VLM returns a
softmax $p_{\text{VLM}}(f)\in[0,1]$ over the markers. In parallel, each
frontier inherits the BVM statistics of its anchor cell and yields a
Bayesian Upper-Confidence-Bound score
\begin{equation}
\mathrm{UCB}(f) \;=\; \mu(f) \;+\; \beta\,\sqrt{\sigma^2(f)},
\label{eq:ucb}
\end{equation}
where $\beta\!>\!0$ controls the exploration weight; the resulting score
inflates the value of cells that are believed relevant on average
\emph{or} have not been sufficiently observed. The two signals are
combined through a convex blend
\begin{equation}
s(f) \;=\; (1 - w)\,p_{\text{VLM}}(f) \;+\; w\,\widetilde{\mathrm{UCB}}(f),
\qquad w = 0.50,
\label{eq:blend}
\end{equation}
where $\widetilde{\mathrm{UCB}}$ is min-max normalized over $\mathcal{F}_t$.
Eq.~\eqref{eq:blend} treats the VLM as a high-noise oracle on
\emph{plausibility} and the BVM as a low-noise oracle on \emph{coverage};
neither fully overrules the other. An explicit \emph{is-worth} gate, a
Yes/No VLM prompt with threshold
$\mathrm{Perception\_PR}(f) \!\geq\! 0.50$, vetoes candidates the VLM
considers structurally implausible (e.g.\ a refrigerator frontier inside
an already-explored bathroom).

\subsection{Decentralized swarm coordination}
\label{subsec:method-swarm}

AnyGoal is decentralized in the \emph{planning} sense: no centralized
controller selects frontiers or actions, no inter-agent reward is shared,
and there is no explicit message channel beyond synchronization of the
shared BEV/BVM, which is a fused observation rather than a command
substrate. Allocation proceeds
by \emph{sequential greedy bidding}: at each re-evaluation step, agents
are indexed in a fixed deterministic order; agent $i$ picks
\begin{equation}
f_i^\star \;=\; \arg\max_{f \in \mathcal{F}_t}\;
   s(f) \cdot \prod_{j<i} \,\phi\!\big(d(f, f_j^\star)\big),
\qquad
\phi(d) =
\begin{cases}
0.60 & d < 100\,\text{px} \\
1.00 & \text{otherwise},
\end{cases}
\label{eq:alloc}
\end{equation}
where $d(\cdot,\cdot)$ is BEV pixel distance. The multiplicative penalty
discourages two agents from converging on the same frontier without ever
instantiating a joint optimization problem.

Frontier re-evaluation is gated to every $\Delta = 25$ steps: between
re-evaluations each agent simply follows its FMM plan towards
$f_i^\star$. Together with a $1.5\!\times$ commitment hysteresis, an
agent switches from $f_{\text{prev}}$ to $f_{\text{new}}$ only when
$s(f_{\text{new}})\geq 1.5\,s(f_{\text{prev}})$, this prevents the
high-frequency frontier oscillations that we observed in early prototypes
when the VLM softmax fluctuated between near-tied candidates. The shared
map itself is the only coordination channel: BEV occupancy is element-wise
max-fused across agents, and the BVM is fused by the same Gaussian
product as Eq.~\eqref{eq:gaussian-update}.

\section{Experimental Setup}
\label{sec:setup}

\paragraph{Benchmark and simulator.}
We evaluate on the full \textsc{GOAT-Bench}~\citep{khanna2024goat}
\emph{val\_unseen} split: $360$ photorealistic HM3D
episodes~\citep{ramakrishnan2021habitat}, $5$--$10$ heterogeneous subtasks per
episode ($2{,}669$ subtasks in the primary evaluation), $36$ goal
categories, and three goal modalities (object name, RGB image, free-form
language description) interleaved within a single episode. The maximum
subtask budget is $500$ steps; the success radius is $1\,\mathrm{m}$ from
any instance of the goal. We use Habitat-Lab and
Habitat-Sim~\citep{savva2019habitat,szot2021habitat} (tag
\texttt{challenge-2022}) with the Stretch embodiment ($1.41$\,m
height, $0.17$\,m radius, $1.31$\,m camera height) and
\texttt{allow\_sliding=False}. Each agent has a forward-facing
$360\!\times\!640$ RGB-D sensor at $42^{\circ}$ HFOV plus GPS/compass;
the action space is the discrete $\{\textsc{move\_forward},
\textsc{turn\_left}, \textsc{turn\_right}, \textsc{stop}\}$.

\paragraph{Metrics.}
We report standard \textsc{GOAT-Bench} metrics: \textbf{Subtask~SR}
(\textsc{stop} within $1\,\mathrm{m}$ of any goal instance in $\leq\!500$
steps), \textbf{SPL} (geodesic-weighted), \textbf{SoftSPL}, \textbf{DTG}
(mean distance-to-goal at \textsc{stop}), and episode-perfect
\textbf{Episode~SR}. We additionally split failures into \emph{false stops}
($d_{\text{euc}}\!>\!1\,\mathrm{m}$) and \emph{budget exhaustion} (no
\textsc{stop} within $500$ steps); the latter is critical for the
perception ablation of Section~\ref{subsec:results-ablation}.

\paragraph{Configurations.}
Unless stated otherwise, AnyGoal runs with $N\!=\!2$ homogeneous ground
agents, SAM3 perception, SpaceOM-3B VLM, and CLIP ViT-B/32 as image
matcher. Architectural hyperparameters are fixed at the values introduced
in Section~\ref{sec:method}: $w\!=\!0.50$, $\Delta\!=\!25$ steps,
$1.5\!\times$ commitment hysteresis, $100\,\mathrm{px}$ separation
threshold with $0.6$ penalty, and ``is-worth'' threshold $0.50$; the UCB
coefficient $\beta$ and remaining implementation constants follow the
released codebase.\footnote{Anonymous code release:
\url{https://anonymous.4open.science/r/anygoal-B855/}.} In ablations,
GroundedSAM uses tuned thresholds (box $0.55$, text $0.40$) and
YOLO-World uses confidence threshold $0.30$ (coarse-grid search,
appendix).

\paragraph{Hyperparameter rationale and single-run protocol.}
The map resolution ($0.05\,\mathrm{m}/\text{cell}$, $480\!\times\!480$
grid) lets the $1\,\mathrm{m}$ success radius span $20$ cells, matching
the fine end of the $5$--$10\,\mathrm{cm}$ range in VLFM and
Modular~GOAT; the convex weight $w\!=\!0.50$, hysteresis
$1.5\!\times$, and $100\,\mathrm{px}$ separation penalty are zero-shot
defaults whose provenance is tabulated in
Appendix~\ref{app:hyperparams}. A full \emph{val\_unseen} run takes
2.2--6.0 days on a single RTX~4090, so every ablation cell is a single
full-split rollout under fixed VLM temperatures and deterministic
simulator seeds; a two-seed evaluation of the headline configuration
agrees to within $0.2$~pp on Subtask~SR ($52.4\,\% \pm 0.2$~pp,
Appendix~\ref{app:repro}).

\paragraph{Real-world transfer.}
AnyGoal ports to a heterogeneous Unitree Go1 + wheeled-platform swarm
by replacing the SAM3+SpaceOM cloud stack with an onboard
YOLO-World+MobileSAM+CLIP edge stack on Jetson AGX Orin (full
hardware, mapping, and ROS2 configuration in Appendix~\ref{app:realworld}).

\section{Results and Evaluation}
\label{sec:results}

\subsection{State-of-the-art performance in the strict physical regime}
\label{subsec:results-sota}

Across the 360-episode headline run, successful subtasks resolve in
\textbf{83 steps} on average (17\,\% of the 500-step budget), while
failing subtasks consume \textbf{161 steps} on average; the median
subtask length is 52 steps. This gap confirms that failures are
dominated by premature \textsc{stop} calls rather than incomplete
exploration, consistent with the $80.2\,\%$ false-stop rate in
\S\ref{subsec:results-ablation}.

\textbf{Within the strict physical regime that mirrors deployable robot
hardware} (discrete $0.25\,\mathrm{m}$ forward steps, $42^{\circ}$ HFOV
matching the HelloRobot Stretch, no continuous teleportation), AnyGoal
with $N\!=\!2$ achieves \textbf{52.4\,\%} Subtask~SR at \textbf{12.7\,\%}
SPL on \textsc{GOAT-Bench} \emph{val\_unseen} (Table~\ref{tab:sota}, top
block), outperforming the official Modular~GOAT baseline by
\textbf{+27.5~percentage points} on Subtask~SR, a $\mathbf{+110\,\%}$
relative gain, while also exceeding SenseAct-NN Skill~Chain by
$+22.9$~pp and the strongest prior zero-shot single-agent system,
TANGO~\citep{ziliotto2025tango}, by $+20.3$~pp. Even the single-agent
configuration alone reaches \textbf{41.9\,\%}~SR, exceeding every
strict-regime prior system by $+9.8$ to $+17.0$~pp: the architectural
gain therefore originates from the Gaussian BVM, the convex-blended UCB
ranking, and the ``is-worth'' verification gate rather than from raw
agent count.

The lower block of Table~\ref{tab:sota} lists 3D-Mem~\citep{yang20253d}
under a \emph{relaxed continuous regime} in which the simulator's action
space and sensor footprint are altered; we discuss why those metrics are
not on the same scale as the strict-regime numbers in~\S\ref{sec:discussion}.

\begin{table}[t]
\centering
\small
\caption{\textbf{Zero-shot performance on \textsc{GOAT-Bench}
\emph{val\_unseen}} (360 episodes, full 2{,}669-subtask split),
partitioned by simulation regime. AnyGoal sets a new state-of-the-art
in the \emph{strict physical regime}, beating Modular~GOAT by
$\mathbf{+27.5}$~pp on Subtask~SR. \emph{Relaxed continuous regime}
methods alter the simulator's action and sensor model; their numbers
are not directly comparable (\S\ref{sec:discussion}).}
\label{tab:sota}
\begin{tabular}{l l c c}
\toprule
\textbf{Method} & \textbf{Details} & \textbf{Subtask SR (\%)} & \textbf{SPL (\%)} \\
\midrule
\multicolumn{4}{l}{\textit{\textbf{Strict Physical Regime}}~--~discrete
$0.25\,\mathrm{m}$ actions, $42^{\circ}$ HFOV, no teleportation:} \\
Modular GOAT~\citep{khanna2024goat}
                            & Modular, zero-shot                 & 24.9 & 17.2 \\
SenseAct-NN Skill Chain~\citep{khanna2024goat}
                            & End-to-end RL                      & 29.5 & 11.3 \\
TANGO~\citep{ziliotto2025tango}
                            & Zero-shot, single-agent            & 32.1 & 16.5 \\
\textbf{AnyGoal (N=1, Ours)} & Zero-shot, single agent           & \textbf{41.9} & \textbf{14.4} \\
\textbf{AnyGoal (N=2, Ours)} & Zero-shot, \emph{decentralized}   & \textbf{52.4} & \textbf{12.7} \\
\textbf{AnyGoal (N=3, Ours)}
                             & Zero-shot, \emph{decentralized}   & \textbf{55.6} & \textbf{11.8} \\
\midrule
\multicolumn{4}{l}{\textit{\textbf{Relaxed Continuous Regime}}~--~continuous
pathfinder teleportation, $120^{\circ}$ HFOV (not directly comparable):} \\
3D-Mem (GPT-4o, closed-API)~\citep{yang20253d}
                            & Zero-shot, 3D snapshot memory      & 62.9 & 44.7 \\
\bottomrule
\end{tabular}
\\[1ex]
\end{table}

\subsection{Multi-agent scaling: $N\!=\!1\!\rightarrow\!2\!\rightarrow\!3$}
\label{subsec:results-coop}

\begin{figure}[t]
\centering
\includegraphics[width=0.60\linewidth]{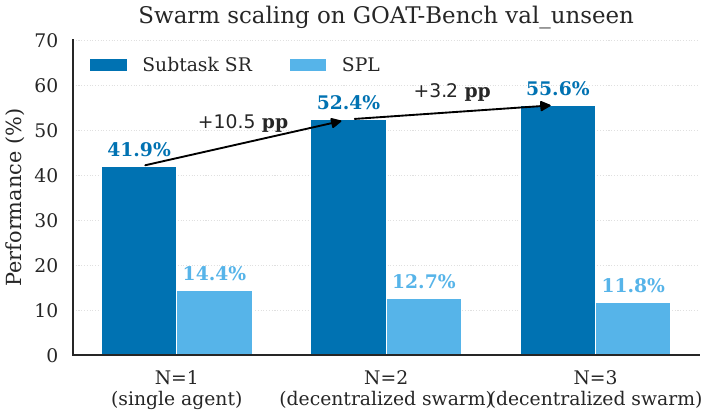}
\caption{\textbf{Swarm scaling on \textsc{GOAT-Bench} \emph{val\_unseen}.}
With the AnyGoal architecture held fixed (SAM3 perception, SpaceOM VLM,
CLIP image matcher, BVM, convex-blended UCB, FMM planner) and only the
number of agents varied, Subtask~SR rises by $+10.5$~pp
($N\!=\!1\!\rightarrow\!2$) and $+3.2$~pp ($N\!=\!2\!\rightarrow\!3$);
SPL declines modestly ($14.4\!\to\!12.7\!\to\!11.8\,\%$), reflecting
extra total trajectory length traded for the SR gain. The $N\!=\!3$
gain is concentrated in description and image-goal modalities, while
object SR is near-saturated at $N\!=\!2$ (see \S\ref{subsec:results-coop}).}
\label{fig:scaling}
\vspace{-2mm}
\end{figure}

Holding the AnyGoal architecture fixed and varying only the number of
agents traces a monotonic scaling curve
(Fig.~\ref{fig:scaling}): Subtask~SR rises $41.9\!\to\!52.4\!\to\!55.6\,\%$
for $N\!=\!1,2,3$. The first step ($+10.5$~pp) is the largest and the
second ($+3.2$~pp) is smaller, consistent with diminishing returns as
frontier supply saturates. The $N\!=\!3$ gain is concentrated in
description and image-goal modalities (Appendix~\ref{app:permodality});
object-name navigation is already near-saturated at $N\!=\!2$ because
open-vocabulary detectors fire reliably on named objects within two
agents' combined field of view.

The architectural lesson is independent of agent count: even at
$N\!=\!1$, AnyGoal exceeds Modular~GOAT, SenseAct-NN Skill~Chain, and
TANGO by $+9.8$ to $+17.0$~pp (Table~\ref{tab:sota}), so the Gaussian
BVM, UCB blend, and ``is-worth'' gate drive most of the gap to prior
art before any multi-agent contribution. The additional $+10.5$ and
$+3.2$~pp at $N\!=\!2,3$ are consistent with later agents mitigating
the single-viewpoint failure modes that bound 3D-Mem-style snapshots:
a hallucination from agent~$i$ with low cone-confidence is rapidly
down-weighted by a high-confidence contradiction from agent~$j$ via
Eq.~\eqref{eq:gaussian-update}, without either modeling the other's
belief explicitly, and \emph{without any centralized controller}, since
only the shared BEV/BVM and the separation penalty
(Eq.~\eqref{eq:alloc}) couple the agents~\citep{burgard2005coordinated,gerkey2002sold}.

\subsection{Perception bottleneck ablation (the 4-way test)}
\label{subsec:results-ablation}

\begin{figure}[t]
    \centering
    \begin{minipage}{0.55\textwidth}
        \centering
        \small
        \captionof{table}{\textbf{Perception backbone ablation on \textsc{GOAT-Bench} \emph{val\_unseen}.} All other components... are held fixed. The \emph{failure-mode shift} is the central scientific finding.}
        \label{tab:ablation}
        \resizebox{\linewidth}{!}{%
        \begin{tabular}{l c c c c}
        \toprule
        \textbf{Perception backbone} & \textbf{Subtask SR (\%)} & \textbf{SPL (\%)} & \textbf{False stop (\%)} & \textbf{Budget exhausted (\%)} \\
        \midrule
        SAM3 + CLIP \textbf{(default)}      & \textbf{52.4} & \textbf{12.7} & 80.2 & 19.8 \\
        SAM3 + DINOv2                        & 50.7 & 13.0 & 79.1 & 20.9 \\
        GroundedSAM~\citep{ren2024grounded}  & 44.8 &  8.5 & 60.7 & 39.3 \\
        YOLO-World~\citep{cheng2024yolo}     & 42.0$^{\flat}$ &  8.4 & 57.7 & 42.3 \\
        \bottomrule
        \end{tabular}%
        }
        \\[0.4ex]
        {\scriptsize $^{\flat}$359/360 ep completed; reported value is $0.420\pm0.001$.}
    \end{minipage}\hfill
    \begin{minipage}{0.42\textwidth}
        \centering
        \includegraphics[width=\linewidth]{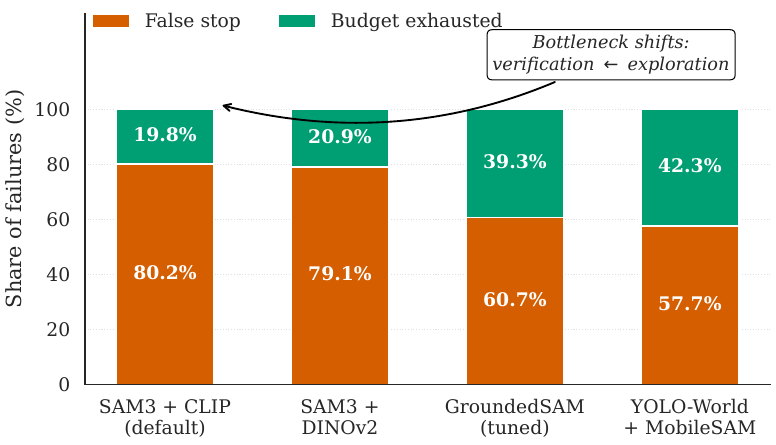}
        \caption{\textbf{Failure-mode shift.} As detector recall increases (right-to-left), budget exhaustion collapses, and false stops dominate.}
        \label{fig:ablation}
    \end{minipage}
\end{figure}

We run the full \textsc{GOAT-Bench} \emph{val\_unseen} pipeline four times
under identical hyperparameters, varying only the perception backbone
(Table~\ref{tab:ablation}). SAM3$+$CLIP achieves the highest
Subtask~SR ($52.4\,\%$); swapping CLIP for DINOv2 trades $1.7$~pp of SR
for marginal SPL gain, GroundedSAM costs $7.6$~pp, and
YOLO-World$+$MobileSAM costs $10.4$~pp. Two-proportion $z$-tests
(Appendix~\ref{app:ztests}) confirm a recall-tier transition: SAM3-class
detectors are significantly stronger than the GroundedSAM/YOLO-World
tier ($p\!<\!10^{-5}$), while within-tier gaps are not significant at
$\alpha\!=\!0.05$.

The deeper finding (right-hand columns of Table~\ref{tab:ablation};
Fig.~\ref{fig:ablation}): as detector recall increases, the failure
composition flips, from $42.3\,\%$ \emph{budget exhaustion} under the
cautious YOLO-World$+$MobileSAM (the detector never returns a valid
mask in $500$ steps) to $80.2\,\%$ \emph{false stops} under SAM3$+$CLIP
(the agent rarely fails to \emph{find} a candidate but fails to
\emph{verify} it). This is, to our knowledge, the first controlled,
like-for-like demonstration that the bottleneck of zero-shot VLM
navigation has shifted: for closed-set systems~\citep{khanna2024goat}
the limiter is detector recall; for SAM3-class open-vocabulary systems
it is \emph{goal verification at the stopping decision}, making the
$80.2\,\%$ false-stop share the primary open
problem~(\S\ref{sec:discussion}).

\section{Discussion}
\label{sec:discussion}

\paragraph{Interpreting SPL across simulation regimes.}
The \textsc{GOAT-Bench} literature has forked into two operating
regimes. The \emph{strict physical regime}
(\citealp{khanna2024goat}; Modular~GOAT, SenseAct-NN,
TANGO~\citep{ziliotto2025tango}, AnyGoal) fixes discrete
$0.25\,\mathrm{m}$ steps, $30^{\circ}$ turns, a $42^{\circ}$ HFOV
(HelloRobot Stretch), and \texttt{allow\_sliding=False}. A
\emph{relaxed continuous regime} used by
3D-Mem~\citep{yang20253d} and DORAEMON~\citep{gu2025doraemon}
substitutes Habitat's oracle pathfinder (collapsing the SPL
trajectory denominator from thousands of steps to a handful of
waypoints) and a $120^{\circ}$ HFOV camera; we read 3D-Mem's
$44.7\,\%$ SPL as a measurement under a different evaluation
contract rather than a comparable upper bound on AnyGoal's
$12.7\,\%$. Inside the strict regime, AnyGoal's SPL exceeds
SenseAct-NN ($12.7$ vs $11.3$); the marginal deficits to Modular~GOAT
and TANGO ($\!-\!4.5$ and $\!-\!3.8$~pp) are the direct cost of the
$+27.5$ and $+20.3$~pp Subtask~SR advantages under a uniform BVM
prior that physically tests more frontiers per subtask.

\section{Limitations and Future Work}
\label{sec:limitations}

\paragraph{Transparent-object floor (perception assumption).}
Twelve goal categories (\emph{glass}, \emph{shower glass},
\emph{handrail}, $\ldots$) score $0\,\%$ SR across all four detector
backbones because the GoalProjector cannot anchor a back-projection on
transparent, thin, or specular surfaces; a DistGate
\texttt{proj\_dist=None} edge case further accepts \textsc{stop} when
no depth is available (one-line patch and a monocular-depth fusion
remedy in Appendix~\ref{app:distgate}).

\paragraph{Odometry and centralized-simulator assumption.}
BVM fusion assumes consistent BEV alignment across agents; the
real-world deployment relies on Vicon ground truth to isolate this
variable, and GPS-denied operation at scale would require onboard
SLAM or UWB-anchored relative localization. The simulator further
shares the BVM as an in-process tensor; multi-machine deployment
would require explicit map synchronization with non-trivial bandwidth
($\sim\!18\,\mathrm{MB}$ per full BEV broadcast).

\paragraph{Single-run evaluation.}
Due to the compute constraints noted in \S4, each ablation cell is a single rollout. A second-seed re-run of the headline configuration is provided (Appendix~\ref{app:repro}), but a multi-seeded sweep across all backbones remains future work.

\paragraph{Path efficiency gap (future work).}
AnyGoal's strict-regime SPL ($12.7\,\%$) trails Modular~GOAT
($17.2\,\%$) and TANGO ($16.5\,\%$) by 3--5~pp because a uniform BVM
prior obliges the agent to physically test more frontiers per subtask.
Two avenues preserve zero-shot generalization while closing this gap:
pre-priming the BVM from a VLM layout caption (``ovens lie near
sinks''), or distilling AnyGoal trajectories into a learned policy
over BVM features while keeping the discrete physical action space.

\section{Conclusion}
\label{sec:conclusion}
\textbf{AnyGoal} is a decentralized, training-free multi-agent
architecture that fuses VLM common sense with a lightweight 2D
Gaussian Bayesian Value Map. On \textsc{GOAT-Bench} \emph{val\_unseen}
it reaches \textbf{52.4\,\%} Subtask~SR (\textbf{+27.5~pp} over
Modular~GOAT) at $N\!=\!2$ and \textbf{41.9\,\%} at $N\!=\!1$, a new
state-of-the-art in the strict physical regime; the gain is
architectural rather than a function of agent count, and under SAM3
$80.2\,\%$ of failures are \emph{false stops}, shifting the open
problem of zero-shot navigation from \emph{finding} the goal to
\emph{verifying} it at the stopping decision.

\clearpage
\acknowledgments{The authors thank the Intelligent Space Robotics
Laboratory at Skoltech for compute access and the open-source maintainers
of Habitat-Sim, SAM3, GroundingDINO, YOLO-World, CLIP, and DINOv2 for
their continued contributions to embodied-AI research. \emph{Code,
configurations, and trajectory logs will be released at the
camera-ready stage.}}

\bibliography{main}

\input{appendix}

\end{document}

%% file: appendix.tex
%
\clearpage
\appendix
\section*{Supplementary Material}

This supplement documents (i) the complete hyperparameter manifest of the
released codebase, (ii) verbatim VLM prompt templates, (iii) the per-agent
decision loop in pseudocode, (iv) extended per-modality, per-failure-mode,
step-distribution, and statistical-test results referenced in the body,
(v) the DistGate edge case underlying the transparent-object failure
floor, (vi) the wall-clock and per-stage VLM compute profile, and
(vii) the real-world heterogeneous-swarm deployment configuration.

\section{Implementation Details}
\label{app:impl}

\subsection{Hyperparameter manifest}
\label{app:hyperparams}

All values below are extracted from the released codebase
(\url{https://anonymous.4open.science/r/anygoal-B855/}) with the
file:line where each constant is defined. ``Provenance'' distinguishes
values inherited from the upstream UCB-frontier reference
implementation from values specific to this work and from values
mandated by the GOAT-Bench protocol.

\begin{table}[h]
\centering
\small
\caption{\textbf{Bayesian Value Map (BVM).}}
\label{tab:hp-bvm}
\begin{tabular}{l l l l}
\toprule
\textbf{Parameter} & \textbf{Value} & \textbf{File:line} & \textbf{Provenance} \\
\midrule
UCB coefficient $\beta$               & 1.7   & arguments.py:278               & inherited \\
Convex blend weight $w$               & 0.50  & arguments.py:274               & this work \\
BVM grid                              & $480\!\times\!480$ cells & bayesian\_value\_map.py:118 & this work \\
Pixels per meter                      & 20.0  & bayesian\_value\_map.py:119    & this work \\
Initial mean $\mu_0$                  & 0.5   & bayesian\_value\_map.py:121    & this work \\
Initial variance $\sigma^2_0$         & 0.5   & bayesian\_value\_map.py:122    & this work \\
Min confidence (lower $\sigma^2$ bound) & 0.25 & bayesian\_value\_map.py:124   & this work \\
Update radius                         & 10 px & arguments.py:276               & this work \\
\bottomrule
\end{tabular}
\end{table}

\begin{table}[h]
\centering
\small
\caption{\textbf{Decentralized allocator.}}
\label{tab:hp-alloc}
\begin{tabular}{l l l l}
\toprule
\textbf{Parameter} & \textbf{Value} & \textbf{File:line} & \textbf{Provenance} \\
\midrule
Min frontier separation               & 100 px (${\sim}5\,\mathrm{m}$) & arguments.py:296 & this work \\
Separation penalty weight             & 0.6   & arguments.py:298               & this work \\
Commitment hysteresis multiplier      & 1.5   & arguments.py:354               & this work \\
Re-identify match radius              & 30 px (${\sim}1.5\,\mathrm{m}$) & main.py:2804--2832 & this work \\
Frontier re-evaluation interval $\Delta$ & 25 local steps & arguments.py:147 & this work \\
\bottomrule
\end{tabular}
\end{table}

\begin{table}[h]
\centering
\small
\caption{\textbf{Navigation and embodiment} (Habitat / GOAT-Bench protocol).}
\label{tab:hp-nav}
\begin{tabular}{l l l l}
\toprule
\textbf{Parameter} & \textbf{Value} & \textbf{File:line} & \textbf{Provenance} \\
\midrule
Map resolution                        & $0.05\,\mathrm{m}$/cell & arguments.py:157   & this work \\
Map size                              & $24\!\times\!24\,\mathrm{m}$ & arguments.py:159 & this work \\
HFOV                                  & $42^{\circ}$ & run override        & GOAT-Bench protocol \\
Frame size (W$\times$H)              & $360\!\times\!640$ & arguments.py & GOAT-Bench protocol \\
Camera height                         & $1.31\,\mathrm{m}$ & run Namespace  & HelloRobot Stretch \\
Forward step                          & $0.25\,\mathrm{m}$ & GOAT YAML       & GOAT-Bench protocol \\
Turn angle                            & $30^{\circ}$ & arguments.py:109     & GOAT-Bench protocol \\
Max steps per subtask                 & 500 & arguments.py:119               & GOAT-Bench protocol \\
Success distance                      & $1.0\,\mathrm{m}$ & arguments.py:121 & GOAT-Bench protocol \\
Detector call interval                & every 3 steps & run override         & this work \\
Vision range                          & 100 cells & run Namespace            & this work \\
Frontier spacing                      & $1.5\,\mathrm{m}$ & arguments.py:235 & this work \\
\bottomrule
\end{tabular}
\end{table}

\begin{table}[h]
\centering
\small
\caption{\textbf{Perception backbones (Section~\ref{subsec:results-ablation}).}}
\label{tab:hp-percep}
\begin{tabular}{l l l l}
\toprule
\textbf{Parameter} & \textbf{Value} & \textbf{File:line} & \textbf{Provenance} \\
\midrule
SAM3 min goal-detection confidence    & 0.30  & arguments.py:319 & this work \\
SAM3 server transport                 & ZMQ REQ/REP, pickle, port 5555 & sam3\_server.py:45 & this work \\
GroundingDINO box threshold (default / tuned) & 0.35 / 0.55 & arguments.py:53 & this work \\
GroundingDINO text threshold (default / tuned) & 0.25 / 0.40 & arguments.py:55 & this work \\
SAM checkpoint                        & vit\_h & run Namespace & inherited \\
YOLO-World COCO confidence            & 0.55 & arguments.py:57 & this work \\
YOLO-World text classification        & 0.55 & arguments.py:174 & this work \\
\bottomrule
\end{tabular}
\end{table}

\begin{table}[h]
\centering
\small
\caption{\textbf{Vision-Language Model and goal projection.}}
\label{tab:hp-vlm}
\begin{tabular}{l l l l}
\toprule
\textbf{Parameter} & \textbf{Value} & \textbf{File:line} & \textbf{Provenance} \\
\midrule
VLM HF identifier                     & \texttt{remyxai/SpaceOm} & VLM/spaceom.py:35 & this work \\
VLM weights dtype                     & bfloat16 & VLM/spaceom.py:64 & this work \\
Server transport                      & HTTP, port 8008 & run Namespace & this work \\
Max new tokens (classification)       & 8     & VLM/spaceom.py:255 & this work \\
Max new tokens (free generation)      & 512   & VLM/spaceom.py:122 & this work \\
``Is-worth'' gate threshold           & 0.50  & SystemPrompt.py:682 & this work \\
BLIP2 ITM fallback threshold          & 0.30  & arguments.py:313 & inherited \\
\midrule
GoalProjector confidence threshold    & 0.30  & run Namespace & this work \\
GoalProjector decay rate              & 0.6   & run Namespace & this work \\
GoalProjector map threshold           & 0.5   & run Namespace & this work \\
DistGate max STOP distance            & $3.0\,\mathrm{m}$ & run Namespace & this work \\
\bottomrule
\end{tabular}
\end{table}

\begin{table}[h]
\centering
\small
\caption{\textbf{FMM planner.}}
\label{tab:hp-fmm}
\begin{tabular}{l l l l}
\toprule
\textbf{Parameter} & \textbf{Value} & \textbf{File:line} & \textbf{Provenance} \\
\midrule
Step size                             & 5 cells & utils/fmm\_planner.py:40 & inherited \\
Stop threshold                        & $< 3$ cells (15 cm) & utils/fmm\_planner.py:117 & inherited \\
Traversible dilation kernel           & $7\!\times\!7$ MORPH\_RECT & utils/fmm\_planner.py:56 & inherited \\
\bottomrule
\end{tabular}
\end{table}

\subsection{Map representation and in-process fusion}
\label{app:mapfusion}

AnyGoal runs as a single Python process; the $N$ agents share GPU
memory and there is no inter-process map synchronization in the
simulator. Each agent maintains an individual semantic map tensor
$\texttt{full\_map}[j] \in \mathbb{R}^{20\times 480\times 480}$ with
the following channel allocation:
\begin{center}
\begin{tabular}{l l}
\toprule
\textbf{Channel} & \textbf{Content} \\
\midrule
0       & Obstacle occupancy \\
1       & Explored mask \\
2       & Current agent position \\
3       & Visited mask \\
4--19   & Semantic category one-hots \\
\bottomrule
\end{tabular}
\end{center}

Every global step the agent maps are merged by element-wise max,
\verb|full_map_pred, _ = torch.max(full_map2, 0)| (\texttt{main.py:2285}),
yielding a conservative union BEV used immediately for frontier
extraction (\texttt{Frontiers()}, \texttt{main.py:2286}) and BVM
scoring.
A single \texttt{BayesianValueMap} object is instantiated once and
shared across all agents (not per-agent); any agent that detects the
goal at confidence $\geq\!0.30$ triggers
\texttt{update\_value\_map()} (\texttt{vlm\_agents.py:561--585}), which
writes directly into the shared map. The real-world deployment
(\S\ref{app:realworld}) replaces this with a ROS2 occupancy-grid
broadcast at 2\,Hz.

\section{VLM Prompt Templates}
\label{app:prompts}

All prompts are in \texttt{src/SystemPrompt.py}. Placeholders are shown
in braces; the GOAT-context header (subtask index, current goal,
modality, remaining budget) is prepended verbatim to every prompt at
\texttt{SystemPrompt.py:563--591}.

\subsection{Scene caption (\texttt{Stage1-Caption})}
\label{app:prompt-caption}

\begin{quote}\small\itshape
You are guiding a robot to find a target object indoors. Describe the
scene BRIEFLY:

(1) Room type (kitchen / bedroom / living room / bathroom / hallway /
stairwell / other).
(2) Key objects seen (name, direction, approximate distance in meters).
(3) Openings (doors, corridors, stairs) and which directions are
unexplored.
(4) Target hint: is the target visible? If not, does the room type or
visible objects suggest it could be nearby?

Be concise --- 3--5 sentences maximum.
\end{quote}

\subsection{Is-worth gate (\texttt{Stage3-Gate})}
\label{app:prompt-gate}

System prompt (\texttt{SystemPrompt.py:426--437}):

\begin{quote}\small\itshape
You are deciding whether the robot should keep exploring THIS direction
to find a target object.

If the detector found the target with confidence ${>}80\%$: answer
\textbf{Yes}.
If the room type matches the target's typical location (e.g.\ stove
$\rightarrow$ kitchen, bed $\rightarrow$ bedroom, staircase handrail
$\rightarrow$ hallway/stairwell): answer \textbf{Yes}.
If co-located objects are visible (e.g.\ microwave near stove,
nightstand near bed): answer \textbf{Yes}.
If the room clearly cannot contain the target (e.g.\ looking for a
staircase handrail in a bedroom): answer \textbf{No}.
Ignore generic objects (doors, walls, light switches).

Output ONLY: Yes or No. Nothing else.
\end{quote}

User template (\texttt{SystemPrompt.py:622--630}):
\begin{quote}\small\ttfamily
- Target of navigation: \{TARGET\}\\
- Common-sense placement: A "\{TARGET\}" is typically found
\{PLACEMENT\_HINT\}.\\
- Scene object (Object Detection): \{OBJECT\_DETECTION\}\\
Decision:
\end{quote}

The Yes/No response is parsed into a probability
$\mathrm{Perception\_PR}\in[0,1]$ via first-token logit extraction
(max\_new\_tokens$\,=\,$8). The downstream gate at
\texttt{SystemPrompt.py:682} applies threshold $0.50$.

\subsection{Frontier vs.\ history (\texttt{Stage-FN})}
\label{app:prompt-fn}

System prompt (\texttt{SystemPrompt.py:441--467}), summarized:
the VLM is shown a top-down semantic map with frontier candidates
(black dots, uppercase letters A--D), historical observation points
(green dots, lowercase letters a--d), and the robot's pose, and asked
to answer Yes (explore a new frontier) or No (revisit a historical
point). Decision rules emphasize distance efficiency for SPL, room-type
priors, exploration--exploitation balance, collision-trap detection,
and directional continuity.

\subsection{Frontier ABCD selection (\texttt{Stage4-Select})}
\label{app:prompt-select}

System prompt (\texttt{SystemPrompt.py:470--517}). The VLM is given
the same top-down semantic map plus the robot's egocentric view; for
each labeled frontier it estimates (i) distance, (ii) target
probability under a common-sense placement prior, (iii) an efficiency
score $=$ target probability $/$ distance. The output is constrained
to a single letter $\in\{$A, B, C, D$\}$, which the parser converts to
the softmax over markers. The historical-point variant
(\texttt{SystemPrompt.py:523--551}) follows the same template with
lowercase markers.

\section{Per-Agent Decision Loop}
\label{app:algo}

\begin{figure}[h]
\centering
\fbox{\begin{minipage}{0.96\linewidth}\small
\textbf{Algorithm 1: AnyGoal -- one decision step per agent $i$.}\\[2pt]
\textit{Inputs:} RGB-D observation $o_t^{(i)}$, current goal token
$g_t$, shared BEV $M$, shared BVM $\mathcal{B}=(\mu, \sigma^2)$,
previous frontier $f_{\text{prev}}^{(i)}$, peers' selections
$\{f_j^{\star}\}_{j<i}$.\\
\textit{Hyperparameters:} $\beta$ (UCB), $w$ (blend), $D_{\text{sep}}$
(separation radius), $\phi$ (separation factor), $\eta=1.5$ (hysteresis),
$\Delta=25$ (re-eval interval), $\tau_{\text{worth}}=0.50$.
\begin{enumerate}[leftmargin=1.6em,topsep=2pt,itemsep=1pt]
\item \textit{Perception.} Run open-vocabulary detector on
$o_t^{(i)}$ for goal $g_t$; back-project hits through depth-cone mask
into the shared BEV.
\item \textit{BVM update.} For each visible cell $p$ with
$(\mu_{\text{obs}}, \sigma^2_{\text{obs}})$, apply
Eq.~\eqref{eq:gaussian-update}.
\item \textit{Merge.} $M \leftarrow \max_j \texttt{full\_map}[j]$
(element-wise across agents).
\item \textit{Frontiers.} Extract $\mathcal{F}_t$ from
connected-component boundary of $M$.
\item \textit{If $t \bmod \Delta = 0$:} run the VLM prompt chain
(scene caption $\to$ is-worth gate $\to$ frontier softmax) to obtain
$p_{\text{VLM}}(f)$ and the gate score
$\mathrm{Perception\_PR}\in[0,1]$. If
$\mathrm{Perception\_PR} < \tau_{\text{worth}}$, set
$p_{\text{VLM}}(\cdot) \leftarrow 0$.
\item \textit{Score.} For each $f \in \mathcal{F}_t$,
$U(f) \leftarrow \mu(f) + \beta\sqrt{\sigma^2(f)}$;
$s(f) \leftarrow (1-w)\,p_{\text{VLM}}(f) + w\,\widetilde{U}(f)$,
with $\widetilde{U}$ min-max normalized over $\mathcal{F}_t$.
\item \textit{Allocate.} $f_i^{\star} \leftarrow \arg\max_f
s(f)\cdot\prod_{j<i}\phi(d(f, f_j^{\star}))$
with $\phi(d){=}0.6$ if $d{<}D_{\text{sep}}$ else $1.0$.
\item \textit{Hysteresis.} Switch from $f_{\text{prev}}^{(i)}$ to
$f_i^{\star}$ only if
$s(f_i^{\star}) \geq \eta\,s(f_{\text{prev}}^{(i)})$;
otherwise hold $f_{\text{prev}}^{(i)}$.
\item \textit{Plan and act.} FMM plan from current pose to
$f_i^{\star}$; execute one discrete action in
$\{\textsc{forward}, \textsc{left}, \textsc{right}, \textsc{stop}\}$.
\item \textit{DistGate STOP check.} If \textsc{stop} is selected,
reject it if the goal projection distance exceeds $3.0\,\mathrm{m}$
(\S\ref{app:distgate}); otherwise emit \textsc{stop}.
\end{enumerate}
\end{minipage}}
\label{alg:anygoal}
\end{figure}

\noindent
Source pointers for each step: BVM update in
\texttt{mapping/bayesian\_value\_map.py:219};
score blend in \texttt{main.py:2735--2738}; separation penalty in
\texttt{main.py:2746--2759}; hysteresis in \texttt{main.py:2804--2832};
DistGate in \texttt{main.py:3239--3253}.

\section{Extended Results}
\label{app:results}

\subsection{Per-modality $\times$ per-$N$ subtask success rate}
\label{app:permodality}

\begin{table}[h]
\centering
\small
\caption{\textbf{Subtask success rate by goal modality and agent count
(\textsc{GOAT-Bench} \emph{val\_unseen}).} Object-goal navigation
saturates near $N\!=\!2$; description and image modalities continue to
benefit from a third agent.}
\label{tab:permodality}
\begin{tabular}{l c c c}
\toprule
\textbf{Modality} & \textbf{$N\!=\!1$} & \textbf{$N\!=\!2$ (headline)} & \textbf{$N\!=\!3$} \\
\midrule
Object ($n=991$)       & 54.3 & 66.0 & 63.1 \\
Image ($n=822$)        & 37.0 & 46.4 & 53.0 \\
Description ($n=856$)  & 34.5 & 44.0 & 51.3 \\
\bottomrule
\end{tabular}
\end{table}

\subsection{Per-failure-mode breakdown across detector backbones}
\label{app:failmodes}

Failure counts are sourced verbatim from the run-specific analysis
files (\texttt{goat\_final\_results\_20260414.md} for SAM3+CLIP;
\texttt{goat\_final\_results\_20260421.md} for SAM3+DINOv2;
\texttt{goat\_gsam\_tuned\_results\_20260422.md} for GroundedSAM;
\texttt{goat\_yw\_results\_20260430.md} for YOLO-World).

\begin{table}[h]
\centering
\small
\caption{\textbf{Failure-mode tally across detector backbones} on
\textsc{GOAT-Bench} \emph{val\_unseen} ($n_{\text{total}}{=}2{,}669$
subtasks; YOLO-World 2{,}667 due to one episode lost to a machine
restart).}
\label{tab:failmodes-full}
\begin{tabular}{l c c c c c}
\toprule
\textbf{Detector} & \textbf{$n_{\text{succ}}$} & \textbf{False stop} & \textbf{Budget exh.} & \textbf{FS share} & \textbf{BE share} \\
\midrule
SAM3 + CLIP \textbf{(default)}     & 1412 & 1008 & 249 & 80.2\,\% & 19.8\,\% \\
SAM3 + DINOv2                      & 1370 & 1028 & 271 & 79.1\,\% & 20.9\,\% \\
GroundedSAM (tuned)                & 1202 &  890 & 577 & 60.7\,\% & 39.3\,\% \\
YOLO-World + MobileSAM             & 1134 &  884 & 649 & 57.7\,\% & 42.3\,\% \\
\bottomrule
\end{tabular}
\end{table}

\subsection{Step-count distribution (headline run)}
\label{app:stepcount}

Parsed from \texttt{goat\_full\_val\_unseen\_20260408.log} per-subtask
termination records (2-agent SAM3+CLIP, 360 episodes, 2{,}669
subtasks).

\begin{table}[h]
\centering
\small
\caption{\textbf{Step-count distribution} on the headline run, by
termination outcome.}
\label{tab:stepcount}
\begin{tabular}{l c c c c c c c c}
\toprule
\textbf{Cohort} & \textbf{$n$} & \textbf{mean} & \textbf{median} & \textbf{P10} & \textbf{P25} & \textbf{P75} & \textbf{P90} & \textbf{max} \\
\midrule
All subtasks          & 2669 & 120.0 & 52  & 10  & 22  & 140 & 433 & 500 \\
Succeeded             & 1412 &  83.4 & 47  &  8  & 21  & 111 & 213 & 497 \\
Failed (combined)     & 1257 & 161.1 & 500 & --- & --- & --- & --- & 500 \\
\quad False stop      & 1008 &  77.4 & 41  & 10  & --- & --- & 191 & 499 \\
\quad Budget exhausted &  249 & 500.0 & 500 & 500 & 500 & 500 & 500 & 500 \\
\bottomrule
\end{tabular}
\end{table}

The bi-modal failure cohort (false stops cluster near 41 steps; budget
exhaustions sit at the 500-step ceiling) substantiates the body claim
that, under a high-recall detector, the dominant failure mode is
premature \textsc{stop} rather than incomplete exploration.

\subsection{Two-proportion $z$-tests for Table~\ref{tab:ablation}}
\label{app:ztests}

Computed from the verified $(n_{\text{total}}, n_{\text{succ}})$ counts
above using the pooled-variance two-proportion $z$-test
($z = (\hat p_1 - \hat p_2) / \sqrt{\hat p (1-\hat p)(1/n_1 + 1/n_2)}$;
$\hat p$ pooled).

\begin{table}[h]
\centering
\small
\caption{\textbf{Two-proportion $z$-tests on detector-ablation Subtask
SR differences.} The recall-tier transition (rows 2, 4, 5) is
significant at $p\!<\!10^{-5}$; within-tier gaps (rows 1, 3) are not
significant at $\alpha\!=\!0.05$.}
\label{tab:ztests}
\begin{tabular}{l c c c c c}
\toprule
\textbf{Pair} & $\hat p_1$ & $\hat p_2$ & $z$ & $p$ (two-sided) & sig \\
\midrule
P1: SAM3+CLIP vs SAM3+DINOv2     & 0.5290 & 0.5133 & $+1.151$ & $0.250$   & ns  \\
P2: SAM3+DINOv2 vs GroundedSAM   & 0.5133 & 0.4504 & $+4.602$ & $4.2\!\times\!10^{-6}$ & *** \\
P3: GroundedSAM vs YOLO-World    & 0.4504 & 0.4252 & $+1.852$ & $0.064$   & ns  \\
P4: SAM3+CLIP vs GroundedSAM     & 0.5290 & 0.4504 & $+5.750$ & $8.9\!\times\!10^{-9}$ & *** \\
P5: SAM3+CLIP vs YOLO-World      & 0.5290 & 0.4252 & $+7.593$ & $3.1\!\times\!10^{-14}$ & *** \\
\bottomrule
\end{tabular}
\\[0.4ex]
{\footnotesize Legend: *** $p\!<\!0.001$,
$\;$ns $p\!\geq\!0.05$. $\hat p$ is the subtask-level success fraction
($n_{\text{succ}}/n_{\text{total}}$); the small gap to the
episode-averaged \textsc{goat\_SR} in Table~\ref{tab:ablation}
($\leq\!0.005$) does not affect any significance conclusion.}
\end{table}

\subsection{Headline-run reproducibility}
\label{app:repro}

The headline $N\!=\!2$ SAM3+CLIP configuration was re-run on a second
simulator seed over the full $360$-episode split. The two seeds are
effectively identical (Table~\ref{tab:seeds}): Subtask~SR differs by
$0.29$~pp and SPL by $0.01$~pp. We therefore report the headline result
as $52.4\,\% \pm 0.2$~pp Subtask~SR over two seeds, and use seed~1
throughout the main paper.

\begin{table}[h]
\centering
\small
\caption{\textbf{Two-seed reproducibility} of the headline $N\!=\!2$
SAM3+CLIP configuration on the full \textsc{GOAT-Bench}
\emph{val\_unseen} split.}
\label{tab:seeds}
\begin{tabular}{l c c c}
\toprule
\textbf{Metric} & \textbf{Seed 1} & \textbf{Seed 2} & \textbf{Spread} \\
\midrule
Subtask SR (\%) & 52.42 & 52.13 & 0.29 \\
SPL (\%)        & 12.66 & 12.67 & 0.01 \\
DTG (m)         & 2.119 & 2.129 & 0.010 \\
\bottomrule
\end{tabular}
\end{table}

\section{The Transparent-Object Floor and DistGate}
\label{app:distgate}

Twelve goal categories score $0\,\%$ subtask SR across all four
detector backbones: \emph{glass} (0/44), \emph{shower glass} (0/13),
\emph{book} (0/45), \emph{boiler} (0/12), \emph{christmas tree} (0/18),
\emph{footrest} (0/10), \emph{hanging clothes} (0/46), \emph{photo}
(0/9), and four other context-dependent categories
(\emph{parapet}, \emph{handrail}, \emph{stair}, \emph{exercise bike}).
Two failure paths converge on these categories.

\textbf{Depth unreliability.} Transparent (\emph{glass},
\emph{shower glass}), reflective, and thin-flat (\emph{photo},
\emph{calendar}) goals are detected in RGB by the VLM but the depth
channel is absent or noisy, so the GoalProjector cannot back-project
the detection to a stable BEV cell. Without a depth anchor the BVM
is never updated for these categories, leaving them invisible to the
frontier planner.

\textbf{DistGate edge case.} The current STOP gate
(\texttt{main.py:3239--3253}) rejects \textsc{stop} when the goal
projection distance exceeds $\texttt{max\_stop\_dist}=3.0\,\mathrm{m}$,
but logs and \emph{accepts} \textsc{stop} when no projection is
available at all:

\begin{quote}\small\ttfamily
elif pdist is None:\\
\quad logging.info(f"[DistGate] Agent\_\{ai\}: STOP requested but "\\
\quad\quad\quad\quad\quad\quad f"no proj\_dist available (GoalProjector didn't detect).")\\
\quad \# action[ai] remains 0 (STOP) --- NOT overridden
\end{quote}

\noindent
A one-line patch (\verb|action[ai] = 1|) converts the no-projection
branch from a logging hole into a hard STOP rejection, requiring depth
evidence at every stop:

\begin{quote}\small\ttfamily
elif pdist is None:\\
\quad logging.info(f"[DistGate] Agent\_\{ai\}: STOP requested but "\\
\quad\quad\quad\quad\quad\quad f"no proj\_dist available --- STOP rejected.")\\
\quad action[ai] = 1  \# override: no depth, no stop
\end{quote}

\noindent
Categories expected to recover under the patch are those that fail
through under-projected detections rather than truly invisible RGB
signals (e.g.\ \emph{photo}, where the GroundedSAM run logs $9$ false
stops at $4.3\,\mathrm{m}$ average euclidean distance to goal). Glass
and shower-glass categories, which fail through absent RGB detections,
remain inaccessible to this patch and motivate the monocular-depth
fusion line of future work in~\S\ref{sec:limitations}.

\section{Compute Profile}
\label{app:compute}

\paragraph{Wall clock.}
The headline run
(\texttt{goat\_full\_val\_unseen\_20260408.log}, 2-agent SAM3+CLIP,
seed~1, 360 episodes, $320{,}247$ global timesteps) terminated at
$6\,\text{d}\,01\,\text{h}\,50\,\text{m}\,53\,\text{s}$
$\approx\!145.8$ wall-clock hours on a single NVIDIA RTX~4090
($\geq\!16$ GB VRAM peak).

\paragraph{Per-stage VLM latency.}
The \texttt{[VLM-LAT]} instrumentation tag was added after the headline
run, so per-stage latency below is measured on the seed~2 re-run
(\texttt{goat\_sam3\_clip\_2agent\_seed2\_20260521\_205309.log}, same
configuration, $249$ episodes at audit time). The model identifier,
hardware, decoding parameters, and code path are identical; the
numbers therefore represent the headline system but are not the
headline run itself.

\begin{table}[h]
\centering
\small
\caption{\textbf{Per-stage VLM latency} (mean / median / P10 / P90, in
seconds; total in hours over $249$ episodes).}
\label{tab:vlmlat}
\begin{tabular}{l c c c c c c}
\toprule
\textbf{Stage} & \texttt{[VLM-LAT]} tag & $n_{\text{calls}}$ & mean & median & P10--P90 & total \\
\midrule
Scene caption  & \texttt{Stage1-Caption} & $9{,}274$ & 0.974 & 0.940 & 0.79--1.15 & 2.51\,h \\
Is-worth gate  & \texttt{Stage3-Gate}    & $9{,}274$ & 0.154 & 0.140 & 0.13--0.16 & 0.40\,h \\
FN decision    & \texttt{Stage-FN}       & $8{,}784$ & 0.192 & 0.180 & 0.12--0.25 & 0.47\,h \\
Frontier ABCD  & \texttt{Stage4-Select}  & $8{,}516$ & 0.266 & 0.240 & 0.15--0.38 & 0.63\,h \\
\midrule
\textbf{Total VLM} & & & & & & \textbf{4.00\,h} \\
\bottomrule
\end{tabular}
\end{table}

\paragraph{Compute breakdown.}
Extrapolating the seed-2 VLM time to the full $360$-episode budget
yields a per-episode VLM overhead of $\approx\!58\,\mathrm{s}$, totalling
$\approx\!5.8\,\mathrm{h}$, or $\approx\!4\%$ of the $145.8\,\mathrm{h}$
headline wall clock. SAM3 detector inference is logged at the
server-side terminal (\texttt{utils/sam3\_server.py:207}) but is not
written to a file; we therefore report SAM3 timing as unmeasured. The
remaining $\geq\!96\%$ of wall-clock is consumed by the Habitat
simulator, the FMM planner, and tensor-level mapping updates, none of
which are individually profiled in the current code. The seed-2
latency profile is reproducible from the released log files.

\section{Real-World Heterogeneous Deployment}
\label{app:realworld}

\paragraph{Platform.}
The real-world swarm comprises a Unitree Go1 quadruped (single-agent
validation) with the same software stack ported to ROS2 in the
\texttt{hetspace-ros2} repository. Per-robot compute is an NVIDIA
Jetson AGX Orin; the sensor is an Intel RealSense D455 RGB-D camera
streamed at $424\!\times\!240\!@\!15\,\mathrm{fps}$ with GPU-accelerated
GLSL pipeline (\texttt{realsense-ros} patched to disable a depth
rotation filter that produced artefacts at the D455 mounting
orientation). Ground-truth pose is supplied by a Vicon
motion-capture rig at $192.168.50.74$.

\paragraph{Mapping.}
The local occupancy/value map is $12\!\times\!12\,\mathrm{m}$ at
$0.05\,\mathrm{m}$/cell ($240\!\times\!240$ grid, world origin
$[-6.0, -6.0]$), updated at $2\,\mathrm{Hz}$ over depth-downsampled
inputs (effective $212\!\times\!120$ projection). Log-odds occupancy
uses $\ell_{\text{occ}}=0.7$, $\ell_{\text{free}}=-0.2$, clipped to
$[-3.0, 3.5]$ with a publish threshold of $1.5$
(\texttt{mapper\_params.yaml:15--40}).

\paragraph{Onboard perception.}
The heavy SAM3+SpaceOM cloud stack is replaced by a fully onboard
combination of YOLO-World (\texttt{yolov8s-worldv2.pt}) with MobileSAM
(\texttt{mobile\_sam.pt}); YOLO confidence threshold $0.4$, IoU
threshold $0.5$, minimum mask area $100\,\mathrm{px}$
(\texttt{mapper\_params.yaml:29--31, 49--50}).

\paragraph{Navigation.}
The deployed FMM planner uses a $5$-cell step (${\sim}0.25\,\mathrm{m}$,
matching the simulator), $0.32\,\mathrm{m}$ robot radius for the Go1
($290\,\mathrm{mm}$ half-width plus $30\,\mathrm{mm}$ margin), and a
$1.0\,\mathrm{m}$ stop distance identical to the simulator success
radius. Each episode is capped at $300$ discrete actions or
$600\,\mathrm{s}$, whichever comes first. The VLM perception gate
threshold is set to $0.60$ on the real robot
(\texttt{nav\_params.yaml:41}) versus $0.50$ in simulation; the higher
threshold reduces false stops on the noisier onboard detector. The
VLM server runs remotely at $192.168.50.185\!:\!8008$ with a $20\,\text{s}$
HTTP timeout (\texttt{nav\_params.yaml:8}).

\paragraph{Inter-node communication.}
All inter-node traffic uses ROS2 topics and services (not the ZMQ
detector-server channel used in simulation). Key interfaces:
\texttt{/cmd\_vel} (\texttt{geometry\_msgs/Twist}) from controller to
\texttt{unitree\_ros}; \texttt{/cmd/discrete\_action}
(\texttt{hetspace\_discrete\_ctrl/DiscreteCmd}) from navigation to
controller; \texttt{/map/occupancy}
(\texttt{nav\_msgs/OccupancyGrid}) from mapping to navigation;
\texttt{/map/targets} (\texttt{visualization\_msgs/MarkerArray}) from
mapping to navigation; \texttt{/vlm/query}
(\texttt{hetspace\_nav/VLMQuery}) from navigation to the VLM client;
and \texttt{/vicon/unitree/\ldots} (\texttt{geometry\_msgs/PoseStamped}
$+$ \texttt{nav\_msgs/Odometry}) from the Vicon bridge to all nodes.
Per-call VLM latency is logged on the client
(\texttt{vlm\_client\_node.py:70, 95, 103, 109, 122}); aggregated
on-robot timing is intentionally deferred to the camera-ready deployment writeup.